\documentclass[10pt,conference,twocolumns]{IEEEtran}
\usepackage{tikz}
\usepackage{amsmath}
\usepackage{cite}
\usepackage{graphicx}
\usepackage{amsfonts}
\usepackage{latexsym}
\usepackage{subfigure}
\usepackage{algorithm}
\usepackage{color}
\usepackage{times}
\usepackage{epsfig, graphics}
\usepackage{bm}
\usepackage{subfigure}
\usepackage{graphicx,wrapfig,lipsum}
\usepackage{epstopdf}
\usepackage{booktabs}
\begin{document}

\newtheorem{theorem}{Theorem}
\newtheorem{lemma}{Lemma}
\newtheorem{conjecture}{Conjecture}
\newtheorem{corollary}{Corollary}
\newtheorem{definition}{Definition}
\newtheorem{scheme}{Scheme}
\newcommand{\argmax}{\arg\!\max}
\newcommand{\rev}[1]{{\color{red}#1}} 
\newcommand{\pound}{\operatornamewithlimits{\gtrless}}

\def\layersep{1.85cm}
\IEEEoverridecommandlockouts

\title{Active Deep Learning Attacks under Strict Rate Limitations for Online API Calls}
\vspace{-0.75cm}
\author{\IEEEauthorblockN{Yi Shi, Yalin E. Sagduyu, Kemal Davaslioglu, and Jason H. Li}
\vspace{-0.5cm}
\IEEEauthorblockA{\\Intelligent Automation, Inc. \\ Rockville, MD 20855, USA\\
Email:\{yshi, ysagduyu, kdavaslioglu, jli\}@i-a-i.com}
}

\maketitle

\begin{abstract}
Machine learning has been applied to a broad range of applications and some of them are available online as application programming interfaces (APIs) with either free (trial) or paid subscriptions.
In this paper, we study adversarial machine learning in the form of back-box attacks on online classifier APIs.
We start with a deep learning based exploratory (inference) attack, which aims to build a classifier that can provide similar classification results (labels) as the target classifier.
To minimize the difference between the labels returned by the inferred classifier and the target classifier, we show that the deep learning based exploratory attack requires a large number of labeled training data samples.
These labels can be collected by calling the online API, but usually there is some strict rate limitation on the number of allowed API calls. To mitigate the impact of limited training data, we develop an active learning approach that first builds a classifier based on a small number of API calls and uses this classifier to select samples to further collect their labels.
Then, a new classifier is built using more training data samples.
This updating process can be repeated multiple times.
We show that this active learning approach can build an adversarial classifier with a small statistical difference from the target classifier using only a limited number of training data samples.
We further consider evasion and causative (poisoning) attacks based on the inferred classifier that is built by the exploratory attack. Evasion attack determines samples that the target classifier is likely to misclassify, whereas causative attack provides erroneous training data samples to reduce the reliability of the re-trained classifier.
The success of these attacks show that adversarial machine learning emerges as a feasible threat in the realistic case with limited training data.
\end{abstract}


\section{Introduction}
As a subfield of \emph{artificial intelligence} (AI), \emph{machine learning} has been used in many applications including cyber security, intelligence analysis, Internet of Things (IoT), cyber-physical systems, autonomous driving, and traditional or computer games.
Machine learning provides the automated means to learn with data.
Recently, significant progress has been made with deep neural networks (\emph{deep learning}) supported by advances in hardware and computational capabilities, making machine learning much more effective than ever before.
Some of the machine learning systems are made available to public or paid subscribers via an application programming interface (API), e.g., Google's Vision API provides image content analysis via REST API \cite{GoogleVision}.
A user can call such an API for the specified data and obtain the returned results, which may be further analyzed for different subsequent tasks.

The online service paradigm, although convenient, makes machine learning vulnerable to various attacks.
Machine learning applications may involve sensitive and/or proprietary information that includes training data, machine learning algorithm, hyperparameters, and functionality of underlying tasks.
By making machine learning applications online, in addition to traditional attacks (e.g., DoS attack), these applications are also becoming subject to various new exploits and attacks (such as stealing the application functionality, trained model, and training data).

The field of \emph{adversarial machine learning} has emerged to understand the machine learning behavior in the presence of an \emph{adversary}.
One example is from the image recognition application, where the adversary generates an image by slightly perturbing real images to fool a state-of-the-art image classifier \cite{GoodfellowShlens}.
As a consequence, a perturbed panda image is recognized as a gibbon by machine learning, while the original image is recognized as a panda by a human.
Adversarial machine learning is still a new research field and the security limitations and vulnerabilities of machine learning are not well understood yet. With the increasing use of machine learning in critical commercial and government applications, the security implications ranging from intellectual property protection to national security raise the need to identify effects of adversarial machine learning and provide the foundation for attack mitigation techniques.

There have been efforts to develop various attacks based on adversarial machine learning.
\emph{Exploratory (or inference) attacks} \cite{Barreno, Tramer, bookchapter2018, Papernot3, Shi17:DL, Ateniese,Fredrikson,Tramer17, Yi2018, Tugba2018} aim to understand how the underlying machine learning algorithm works for an application, and consequently infer any \emph{sensitive and/or proprietary information}. An exploratory attack can be launched under different assumptions made on the attacker's knowledge. For instance, a black-box attack can be launched without any prior knowledge on the machine learning algorithm and the training data  \cite{Tramer, bookchapter2018,Papernot3, Shi17:DL}. As shown in \cite{Tramer, bookchapter2018,Papernot3, Shi17:DL}, the adversary can call the target classifier $T$ with a large number of samples, collect the labels on these samples, and then use this data to train a deep learning classifier $\hat T$ as an estimate of the target classifier $T$.
This attack implicitly steals the underlying training data, the inner-workings of the machine learning algorithm, and its hyperparameter selection that constitute the core of intellectual property (which has been often built with long-time research and development investment). Moreover, once $\hat T$ is inferred, further attacks such as \emph{evasion attacks} or \emph{causative (poisoning) attacks} can be launched \cite{ShiMilcom}.


One major assumption of exploratory attacks is that an adversary can collect training data from the target machine learning application. Supervised machine learning relies on the quality and the quantity of training data samples. Without sufficient training data, it may not be possible to train a successful machine learning algorithm \cite{Kemal2018}.
However, machine learning APIs, in general, have \emph{strict rate limitations} on how many API calls can be made in a certain period of time (ranging from per second to per day).
In addition, a user who makes too frequent API calls may be identified as a potential adversary and may be blocked later.
Hence, we consider the practical case that imposes strict rate limitations on adversarial machine learning.
In this paper, we select a real online machine learning API for text analysis \cite{Datumbox} as the target classifier $T$.
The adversary treats $T$ as a black box, since the underlying machine learning algorithm, parameters, and training data are all unknown.

\begin{figure}
\centering
\includegraphics[width=\columnwidth]{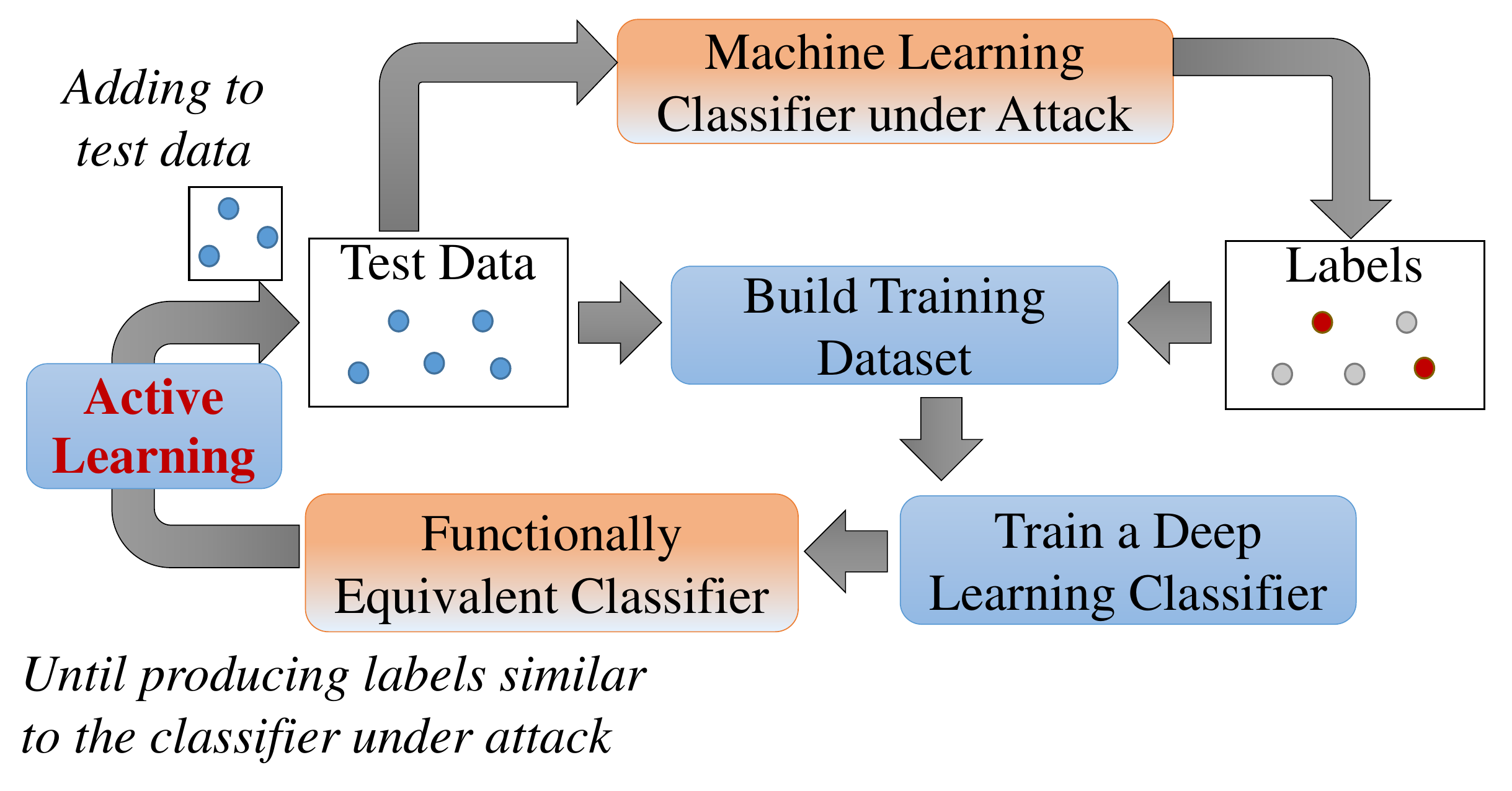}
\caption{Adversarial deep learning with limited training.}\label{fig:systemfigure}
\end{figure}

First, we consider exploratory attacks.
We show that when the number of API calls is limited, the adversary cannot infer the target classifier $T$ with acceptable accuracy, even when it launches a deep learning based exploratory attack.
Thus, we design an approach based on \emph{active learning} \cite{Pi,Settles} to enhance the training process of the exploratory attack. The adversary first builds a classifier $\hat T$ based on a small number of samples and collects their labels.
Using active learning, it further selects additional samples using $\hat T$ and sends only samples with the low classification confidence by $\hat T$ to the target classifier $T$ to be labeled.
These samples are used as additional training data to improve the inferred classifier $\hat T$.
This updating process can be repeated multiple times.
The system model is shown in Figure~\ref{fig:systemfigure}. We show that the proposed method significantly improves the performance of the inferred classifier $\hat T$ that is measured as the statistical difference of classification results, i.e., labels returned by the original classifier $T$ and the inferred classifier $\hat T$. This approach provides insights into situational awareness, attack modeling, detection, and mitigation in different applications of cyberspace \cite{TACT, RADAR, STU}.

Once the exploratory attack is successful, the adversary can launch further attacks such as \emph{evasion} and \emph{causative attacks} by using the inferred classifier $\hat T$.
The evasion attack \cite{Papernot3,bookchapter2018,Biggio,Kurakin, Papernot2} aims to provide the target classifier with some test data that will likely result in incorrect labels (e.g.,  a spam email filter is fooled into accepting spam emails as legitimate). The adversary runs some samples through the inferred classifier $\hat T$ based on a deep neural network and further examines their classification scores provided by the classifier.
If these scores (likelihoods of labels returned by the deep neural network) are within the decision region for label $j$ and are close to the decision region for another label $i$, these sample are selected such that the target classifier $T$ is very likely to misclassify data samples from label $i$ as label $j$.
Such samples can be used to select the test data samples in the form of evasion attacks.


On the other hand, the causative attack \cite{Biggio12, Kermani15, Pi, YiMilcom2018, Alfed16} aims to provide the target classifier with incorrect information as the additional training data samples to reduce the reliability of the re-trained classifier.
For this purpose, the adversary sends some data samples to its inferred classifier $\hat T$.
If the deep learning scores are far away from the decision boundary, the adversary changes their labels and sends these mislabeled samples as additional training data to the target classifier $T$. The target classifier that is re-trained this way is expected to perform significantly worse than before.

The problem of limited training data in adversarial machine learning has been studied in \cite{Papernot3}, where the adversary first infers the target classifier with limited real training data, and then generates synthetic data samples from real data in an evasion attack by adding adversarial perturbations that are gradually improved by the derivative and checking labels queried from the target classifier.
The difference is that we do not use synthetic data in the evasion attack and our goal is to improve the inferred classifier (not the adversarial samples) that can be used later to select an arbitrary number of samples for evasion or causative attacks without further querying the target classifier.

The rest of the paper is organized as follows. Section~\ref{sec:active} studies the exploratory attack assuming that a large number of API calls is allowed and then presents an active learning based approach to launch the exploratory attack under limitations on API calls.
Section~\ref{sec:ceattacks} investigates the subsequent evasion and causative attacks. Section~\ref{sec:conclusion} concludes the paper.

\section{Exploratory Attack and Training Data Refinement with Active Learning}
\label{sec:active}

\subsection{Exploratory Attack on an Online Classifier}
\label{sec:Datumbox}

Many machine learning services are made available online as APIs. Users can use these services by making API calls and integrate their API outputs with their own applications. Such applications include, but not limited to, topic classification, sentiment analysis, subjectivity analysis, keyword extraction, language detection, and spam detection. APIs usually provide free subscriptions that allow a user to call them only few times in a certain time frame.
For example, DatumBox provides a number of machine learning APIs for text analysis \cite{Datumbox}. A user can specify the input text data, call a DatumBox API to analyze this text, and obtain results such as classification of the input text as subjective or objective.

This paradigm raises potential security issues for online services, since an adversary can also access the input and output of these services.
In particular, the adversary can perform a black-box exploratory attack, where the adversary does not have any knowledge on the target classifier $T$, including the training data, training algorithm, and machine learning parameters.
Such an attack can be launched as follows:
\begin{enumerate}
	\item The adversary calls the online API of the target classifier $T$ multiple times using a set $S$ of samples and for each sample $s \in S$, the adversary collects the label $T(s)$ returned by the online API.
	
	\item By using $S$ and the collected labels, the adversary trains a deep neural network to build a classifier $\hat T$, as an estimation of $T$.
\end{enumerate}

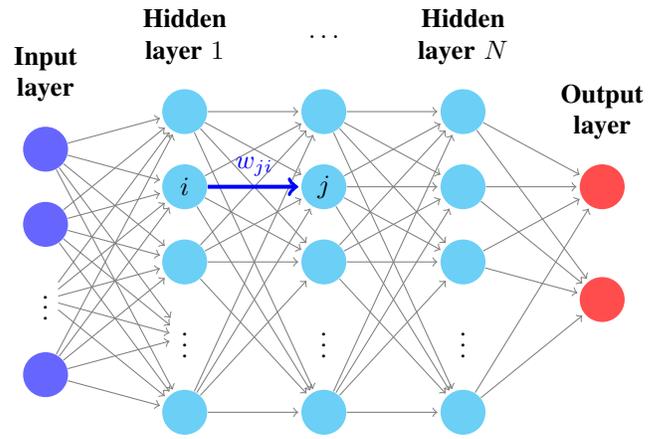
\begin{figure}
	\centering

\begin{tikzpicture}[shorten >=1pt,->,draw=black!50, node distance=\layersep]
\tikzstyle{every pin edge}=[<-,shorten <=1pt]
\tikzstyle{neuron}=[circle,fill=black!25,minimum size=17pt,inner sep=0pt]
\tikzstyle{input neuron}=[neuron, fill=blue!60];
\tikzstyle{output neuron}=[neuron, fill=red!70];
\tikzstyle{hidden neuron}=[neuron, fill=cyan!50];
\tikzstyle{annot} = [text width=4em, text centered]

\foreach \name / \y in {1,2,4}
\node[input neuron] (I-\name) at (0,-\y) {};
\foreach \name / \y in {3}	
\node[draw=none] (I-\name) at (0,-\y) {$\vdots$};

\foreach \name / \y in {1,3,5}
\path[yshift=0.5cm] node[hidden neuron] (H-\name) at (\layersep,-\y cm) {};

\foreach \name / \y in {4}
\path[yshift=0.5cm] node[draw=none] (H-\name) at (\layersep,-\y cm) {$\vdots$};

\foreach \name / \y in {2}
\path[yshift=0.5cm] node[hidden neuron] (H-\name) at (\layersep,-\y cm) {$i$};
\foreach \name / \y in {2}
\path[yshift=0.5cm] node[hidden neuron] (H2-\name) at (2*\layersep,-\y cm) {$j$};			


\foreach \name / \y in {1,3,5}
\path[yshift=0.5cm]
node[hidden neuron] (H2-\name) at (2*\layersep,-\y cm) {};

\foreach \name / \y in {4}
\path[yshift=0.5cm] node[draw=none] (H2-\name) at (2*\layersep,-\y cm) {$\vdots$};

\foreach \name / \y in {1,2,3,5}
\path[yshift=0.5cm]
node[hidden neuron] (H3-\name) at (3*\layersep,-\y cm) {};

\foreach \name / \y in {4}
\path[yshift=0.5cm] node[draw=none] (H3-\name) at (3*\layersep,-\y cm) {$\vdots$};

\foreach \name / \y in {1,2}
\node[output  neuron] (O-\name) at (4*\layersep,-1.5*\y cm) {};

\foreach \source in {1,2,3,4}
\foreach \dest in {1,2,3,4,5}
\path (I-\source) edge (H-\dest);

\foreach \source in {1,2,3,5}
\foreach \dest in {1,2,3,5}
\path (H-\source) edge (H2-\dest);

\foreach \source in {1,2,3,5}
\foreach \dest in {1,2,3,5}
\path (H2-\source) edge (H3-\dest);	

\foreach \source in {1,2,3,5}
\foreach \dest in {1,2}
\path (H3-\source) edge (O-\dest);

\foreach \name / \y in {2}
\draw [line width=1.5,blue] (H-\name) -- (H2-\name) node [midway, above] (TextNode) {$w_{ji}$};

\node[annot,above of=I-1, node distance=1cm,font=\bfseries] (hl) {Input layer};
\node[annot,above of=H-1, node distance=1cm,font=\bfseries] (hl) {Hidden layer $1$};
\node[annot,above of=H2-1, node distance=1cm,font=\bfseries] (hl) {$\ldots$};	
\node[annot,above of=H3-1, node distance=1cm,font=\bfseries] (hl) {Hidden layer $N$};	
\node[annot,above of=O-1, node distance=1cm,font=\bfseries] (hl) {Output layer};	
\end{tikzpicture}

\caption{Feedforward neural network.}\label{fig:FNN}
\end{figure}

In this paper, we consider a \emph{black-box exploratory attack}, i.e., the adversary only has a set $S$ of samples to call the online API and the returned labels. The target classifier $T$ has been trained before. The adversary does not know the training data or the algorithm (e.g., Naive Bayes, Support Vector Machine (SVM), or a more sophisticated neural network algorithm) used to build this classifier.
The data is divided into training data and test data. Using the training data, the adversary applies deep learning (based on a deep neural network) to infer $T$.
Deep learning refers using training data to train a deep neural network (namely, finding the weights in the neural network) for computing labels. In particular, we consider a \emph{feedforward neural network} (FNN) shown in Figure.~\ref{fig:FNN}. We use the backpropagation algorithm to train the deep neural network.
A neural network consists of simple elements called \emph{neurons} and weighted connections, a.k.a. \emph{synapses}, among these neurons. A neuron $j$ performs a basic computation over its input synapses $w_{ji}$ from each neuron $i$ that is connected to $j$, and outputs a single scalar value $y_j$, which can be interpreted as its activation or firing rate. In a \emph{feedforward neural network} (FNN) architecture, neurons are arranged in layers and synapses are connecting neurons in one layer to neurons in the next layer. The activations of neurons in the input layer are set externally while the activations of the hidden layer neurons and output layer neurons are computed as specified above. In particular, the activations of the output layer neurons represent the result of the network's computation.

We consider a real online classifier API, namely the subjectivity analysis API of DatumBox \cite{Datumbox}, as the target classifier $T$.
The adversary calls this API with a number of text samples and collects the returned labels, which may be label $1$ (subjective) or label $2$ (objective).
The free subscription allows $1000$ calls per day, i.e.,  labels for $1000$ samples can be collected per day.

For the numerical results presented in this paper, the adversary collects labels for $10000$ data samples over time. Each data sample is the text of a tweet that has been collected from the public Twitter API. In this section, the adversary uses half of the data samples as training data to train a classifier $\hat T$ and the other half as test data to evaluate its performance $d(\hat T, T)$ of $\hat T$.

Deep learning requires representing each sample as a set of features.
We use top word distribution to build such a set of features as follows. The adversary first obtains a list of words $W=(w_1, w_2, \cdots)$ sorted by their frequencies of occurrence in text, i.e., $p_1 \ge p_2 \ge \cdots$ where $p_i$ is the frequency of occurrence for word $w_i$.
Then the features for a sample of text is the set of numbers of occurrence $o_i$ for each word $w_i$.
Thus, $2000$ features are obtained by considering distributions of top $2000$ words.
The adversary uses these features to train the deep learning classifier $\hat T$.

The adversary builds a deep neural network (an FNN) as the classifier  $\hat T$. To optimize the hyperparameters of the deep neural network, the following two measures are computed. 
The difference $d_1(\hat T, T)$ between $\hat T$ and $T$ on label $1$ is the number of samples with ${\hat T}(s)=2$ and $T(s)=1$ divided by the number of samples with label $1$ in test data.
We can define $d_2(\hat T, T)$ similarly.
Then, we aim to minimize $d_{\max}(\hat T, T) = \max\{d_1(\hat T, T),d_2(\hat T, T)\}$ for test data to balance the effect on both labels.

We implement this process by using the Microsoft Cognitive Toolkit (CNTK) \cite{CNTK} in a Python code to train the FNN with the optimal hyperparameters.
The hyperparatemers of the deep neural network are optimized as follows.
\begin{itemize}
	\item The number of hidden layers is $2$.
	\item The number of neurons per layer is $60$.
	\item The loss function (that is used to measure the difference of the labels returned by the deep neural network from the ground truth labels) is cross entropy.
	\item The activation function in hidden layers is sigmoid.
	\item The activation function in output layer is softmax.
	\item Weights and biases are not initially scaled.
	\item Input values are unit normalized in the first training pass.
	\item The minibatch size is $5$.
	\item The momentum coefficient to update the gradient is $0.9$.
	\item The number of epochs per time slot is $8$.
\end{itemize}

The difference between labels returned by $T$ and $\hat T$ for test data is found as
\begin{eqnarray*}
d_1(\hat T, T) = 12.82\%, & d_2(\hat T, T) = 12.88\%.
\end{eqnarray*}
Thus, we obtain
\begin{eqnarray*}
d_{\max}(\hat T, T) = 12.88\%.
\end{eqnarray*}


However, the above attack requires to collect labels for many samples, which needs to be done over multiple days (due to the rate limit of $1000$ samples per day). Next, we will consider the case when the adversary can perform the exploratory attack with a small number of calls.

\subsection{Active Learning with Limited Training Data}

\begin{figure}
\centering
\subfigure[Step 1.]
{\includegraphics[width=0.75\columnwidth]{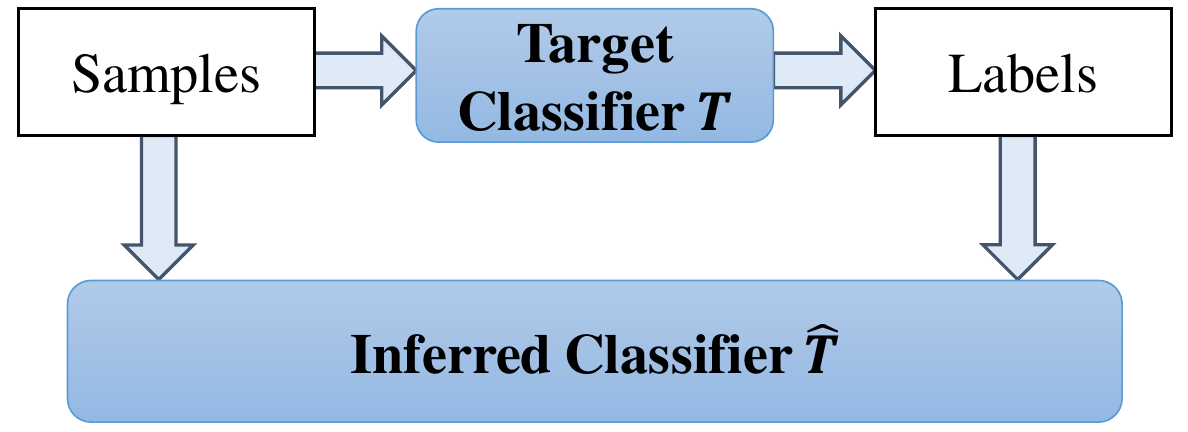}}
\subfigure[Step 2.]
{\includegraphics[width=\columnwidth]{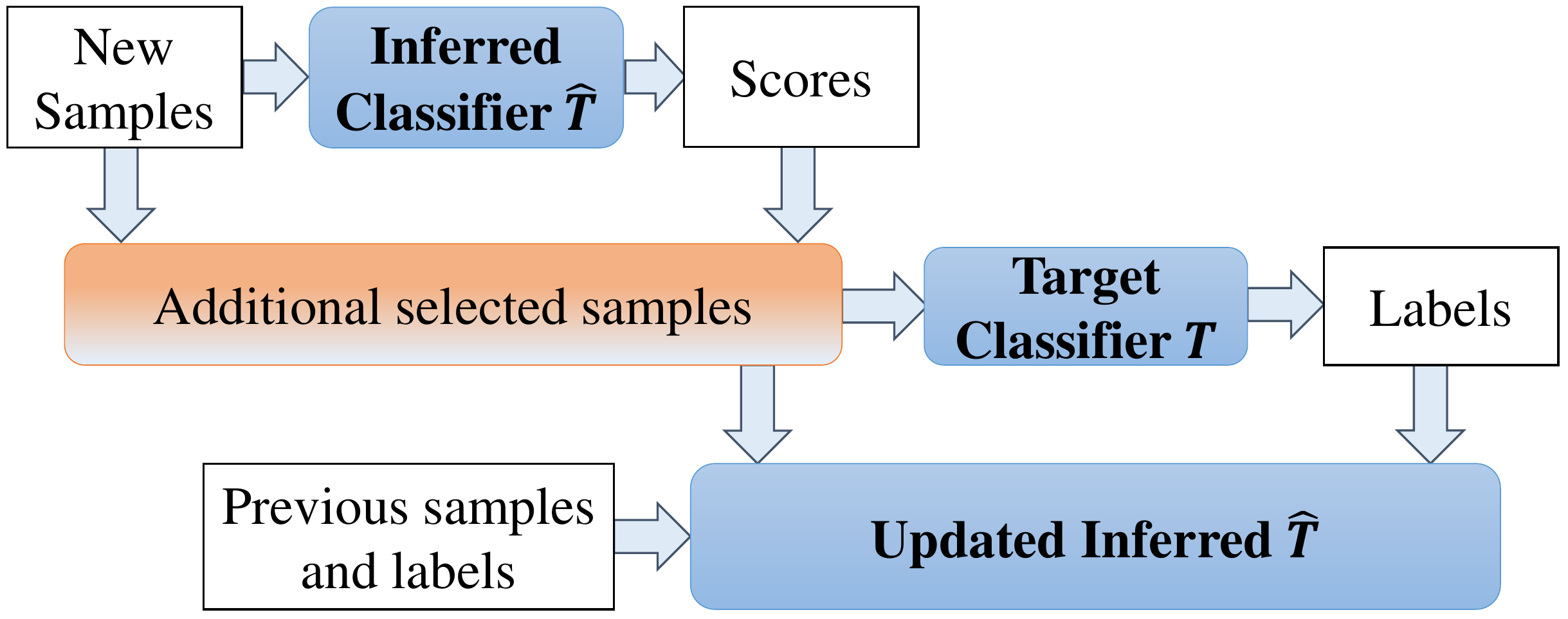}}
\caption{Exploratory attack with active learning.}
\label{fig:active}
\end{figure}

Our approach to mitigate the effect of limited training on the adversary is based on \emph{active learning}, which can jointly and iteratively build the inferred classifier $\hat T$ using limited training data.
Suppose that the adversary starts with 200 samples and their labels obtained from the target classifier $T$ (see Figure~\ref{fig:active}(a)).
Data is split in $100$ training samples (to build the classifier $\hat T$) and $100$ test samples.

Again, the adversary builds the optimal $\hat T$ by selecting deep learning hyperparameters to minimize the difference between $\hat T$ and $T$. These hyperparameters are given by:
\begin{itemize}
	\item The number of hidden layers is $2$.
	\item The number of neurons per layer is $10$.
	\item The loss function is squared error.
	\item The activation function in hidden layers is sigmoid.
	\item The activation function in output layer is softmax.
	\item All weights and biases are initially scaled by $0.5$.
	\item Input values are unit normalized in the first training pass.
	\item The minibatch size is $25$.
	\item The momentum coefficient to update the gradient is $0.9$.
	\item The number of epochs per time slot is $10$.
\end{itemize}

Note that this deep neural network is smaller than the previous one because fewer training data samples are used to build this second neural network.
The difference between labels returned by $T$ and $\hat T$ is found as
\begin{eqnarray*}
d_1(\hat T, T) = 33.78\%, \:\:\:\:\: d_2(\hat T, T) = 30.77\%.
\end{eqnarray*}
Thus, we obtain
\begin{eqnarray*}
	d_{\max}(\hat T, T) = 33.78\%.
\end{eqnarray*}
The adversary then collects more labels from $T$ to improve $\hat T$. In the benchmark approach, randomly selected samples are sent to $T$ and the returned labels are collected and added to the training data in order to improve $\hat T$.

We first take a closer look at the inferred classifier $\hat T$.
For each sample $s$, $\hat T$ first determines a classification likelihood score ${\hat S}(s)$ and then compares it with a threshold $S_{\hat T}$, which is optimized to minimize $d_{\max}(\hat T, T)$. The optimal $S_{\hat T}$ is found as $0.17$.
If ${\hat S}(s) < S_{\hat T}$, sample $s$ is classified as label $1$, otherwise sample $s$ is classified as label $2$. The difference between score ${\hat S}(s)$ and $S_{\hat T}$ captures the confidence of classification. If ${\hat S}(s)$ is close to $S_{\hat T}$, then its is likely that such a classification is wrong.
On the other hand, if ${\hat S}(s)$ is far away from $S_{\hat T}$, then it is unlikely that such a classification is wrong.

\begin{table*}
	\caption{Performance (difference between labels returned by $T$ and $\hat T$) with active learning.}
	\centering
	\small
	\begin{tabular}{c|c|c|c|c|c|c}
		\toprule
		\multicolumn{3}{c}{Training data size} & \multicolumn{2}{|c|}{Active learning} & \multicolumn{2}{|c}{Benchmark} \\ \hline
		Initial Samples & Additional Samples & Total Samples &  $d_1(\hat T, T)$ & $d_2(\hat T, T)$ & $d_1(\hat T, T)$ & $d_2(\hat T, T)$ \\ \hline
		$100$ & $224$ & $324$ & $28.38\%$ & $26.92\%$ & $31.08\%$ & $30.77\%$ \\ \hline
		$100$ & $579$ & $679$ & $24.32\%$ & $23.08\%$ & $29.73\%$ & $30.77\%$ \\ \hline
		$100$ & $1007$ & $1107$ & $18.92\%$ & $19.23\%$ & $31.08\%$ & $30.77\%$ \\ \bottomrule
	\end{tabular}
	\label{tab:active}
\end{table*}

Based on this understanding of ${\hat S}(s)$,  an active learning approach is designed as follows (see Figure~\ref{fig:active}(b)).
\begin{enumerate}
\item The adversary sends randomly selected text samples $s$ to its inferred classifier $\hat T$ and obtains classification likelihood scores ${\hat S}(s)$ from the inferred classifier.

\item If classification likelihood score ${\hat S}(s)$ is close to threshold $S_{\hat T}$, the adversary sends this text sample to classifier $T$ and obtain its label $T(s)$.
The adversary adds text sample $s$ and its label $T(s)$ to the training data.

\item If additional training data is sufficient, the adversary updates $\hat T$. Else, the adversary goes to Step $1$.
\end{enumerate}

Using the above steps, the adversary generates $500$ ($1000$, or $2000$) additional text queries in Step 1 of the training process and obtains the scores of the corresponding samples from the inferred classifier $\hat T$ in Step 2. Based on these scores, $224$ ($579$, or $1007$) text samples close to $S_{\hat T}$ are selected by active learning to call classifier $T$. We refer to these samples as the \emph{actively learned samples}. Using the scores returned by $T$, the inferred $\hat T$ is updated. In the benchmark scheme, we consider that only randomly selected text sample is sent to the original classifier. Active learning performs better than the benchmark because it calls $T$ with data samples that are uncertain with respect to the original inferred classifier $\hat T$, while the benchmark is not able to adapt its decision boundary to the uncertainty of such samples. 
In both active learning and benchmark scheme, the adversary uses the original $100$ test samples to evaluate the performance.

The results are presented in Table~\ref{tab:active}. The total number of training data samples is set the same for active learning and benchmark scheme.
With $224$ additional (actively learned or randomly selected) training data samples, active learning performs better than the benchmark scheme and achieves $d_{\max}(\hat T, T) = 28.38\%$, whereas the benchmark scheme achieves $d_{\max}(\hat T, T)  = 31.08\%$.
This performance gain of active learning further increases with more ($579$ or $1007$) additional training data samples. For example, with 1107 total samples, active learning reduces $d_{\max}(\hat T, T)$ to $19.23\%$ (which corresponds to $38.13\%$ improvement over the benchmark scheme).
Thus, active learning always achieves better performance than the benchmark scheme.
Moreover, with more additional training data samples, the advantage of active learning grows, which shows the importance of identifying and selecting uncertain samples in performance improvement.

\section{Evasion Attacks and Causative Attacks}
\label{sec:ceattacks}

When the adversary completes a successful exploratory attack and infers a classifier $\hat T$ that is similar to the target classifier $T$, it can then try to learn the behavior of $T$ by using $\hat T$ and launch subsequent attacks. In this section, we demonstrate two such attacks, namely, the evasion attack and causative attack.

%

\subsection{Evasion Attack}

The evasion attack aims to find test samples for which labels returned by $T$ are likely wrong.
Although $T$ is unknown by the adversary, the inferred classifier $\hat T$ can used to predict which samples may be incorrectly classified.
There may be different attack objectives and approaches. In one example, the objective is to maximize the average error.
To achieve this objective, the adversary selects samples with classification scores that are close to the threshold (i.e., samples with low confidence on classification). In another example, the objective is to maximize the error of misclassifying a label $1$ sample as label $2$.
To achieve this objective, the adversary selects samples classified with label $2$ and with score that is close to the threshold.
The objective of maximizing the error of misclassifying a label $2$ sample as label $1$ can be achieved similarly.

 We need the ground truth (i.e., the correct label for samples in test data) to evaluate the effectiveness of such attacks.
Since it is assumed that the adversary has no knowledge of ground truth, we omit the performance results for evasion attack.

\subsection{Causative Attack}

To reduce classification errors, some classifiers are updated (re-trained) by user feedback.
That is, a user can review the returned labels by $T$ and manually fix labels on some samples or agree on the returned labels.
If such flags are received, the sample with user defined labels can be used as additional training data to re-train the classifier.
However, such a process also introduces a new type of attack, namely causative attack, where the adversary provides wrong labels such that the updated classifier $\tilde T$ becomes less reliable in terms of classification accuracy.

There are different strategies to provide wrong labels.
In the extreme case, the adversary may provide wrong labels for all data samples to maximize the impact of its attack.
However, it is easy to detect such an attack and block the adversary, i.e., its feedback will not be considered in the re-training process.
Thus, the adversary aims to achieve a significant impact of its attack by using a small number of wrong labels.
Suppose that the adversary can change labels for only $p\%$ of all samples, where $p$ is a small number.
Then, the problem is to determine how to select the best set of data samples and provide wrong labels on this set of data samples.
Clearly, this selection requires the knowledge of $T$, which can be obtained by analyzing $\hat T$ after a successful exploratory attack.

We again consider the classification scores provided by the deep learning based classifier $\hat T$.
If the adversary changes the label for a sample with a score that is far away from the threshold (i.e., with high classification confidence), this new label will be very likely a wrong label and will make the updated classifier $\tilde T$ worse.
On the other hand, if the adversary changes the label for a sample with a score that is close to the threshold (i.e., with low classification confidence), this new label may not be a wrong label and will not make the updated classifier $\tilde T$ worse.
Thus, the adversary performs a causative attack as follows.
\begin{enumerate}
\item The adversary sends some samples to $\hat T$ and receives their scores and labels.
Note that $\hat T$ is the classifier inferred by the adversary and thus there is no rate limitation, i.e., the adversary can collect a large number of samples with scores and labels.

\item The adversary selects samples with top $\frac{p}{2} \%$ scores and samples with bottom $\frac{p}{2} \%$ scores. These scores are far away from the threshold.

\item The adversary switches labels of selected samples and sends these labels as user feedback.
\end{enumerate}
To measure the impact of a causative attack, we compare the outputs of the original classifier $T$ and the updated classifier $\tilde T$.
For samples with label $1$ (by class $T$), suppose that the number of these sample is $n_1$, while $T$ and $\tilde T$ provide different labels on $m_1$ samples.
Then, we define the difference on samples with label $1$ as $d_1(T, \tilde T) = \frac{m_1}{n_1}$.
Similarly, define the difference on samples with label $2$ as $d_2(T, \tilde T) = \frac{m_2}{n_2}$ and the average of the difference on all samples as $d(T, \tilde T) = \frac{m_1+m_2}{n_1+n_2}$, where $n_2$ is the number of samples with label $2$ and $m_2$ is the number of different labels by $T$ and $\tilde T$ on these $n_2$ samples.


We use the classifier $\hat T$ built in Section~\ref{sec:active} with $1107$ training data samples, we apply it on another set of $1000$ samples and perform a causative attack with $p=10$. Then, we obtain
\begin{eqnarray*}
d_1(T, \tilde T) = 39.92\%, & d_2(T, \tilde T) = 55.64\%.
\end{eqnarray*}
Thus, the average difference on all samples is
\begin{eqnarray*}
d(T, \tilde T) = 48.00\%.
\end{eqnarray*}
As a result, the causative attack built upon the exploratory attack reduces the reliability of the updated classifier $\tilde T$ significantly.


\section{Conclusion}
\label{sec:conclusion}

We studied the exploratory attack to infer a target online classifier, when the number of calls to the online API is limited.
Typically, adversarial deep learning requires a large number of training samples. Under strict rate limitations on API calls, we showed that there is a significant difference between the labels returned by the inferred classifier and the target classifier. To mitigate this effect, we designed an active learning process for adversarial deep learning, which first builds a classifier based on limited training data and then uses this classifier to selectively make more API calls.
We showed that this attack can reliably infer the target classifier even with a limited number of API calls. By analyzing the classifier inferred by a successful exploratory attack, the adversary can launch further attacks.
In particular, we designed evasion and causative attacks using the inferred classifier to select test and training data samples, respectively.
The evasion attack identifies test data that the target classifier cannot reliably classify, while the causative attack effectively poisons the re-training process such that the reliability of updated classifier degraded.
We showed that these attacks are feasible with limited training data and adversarial machine learning emerges as a practical threat to online classifier APIs.

\end{document}